\documentclass[10pt,journal,compsoc]{IEEEtran}
\ifCLASSOPTIONcompsoc
  \usepackage[nocompress]{cite}
\else
  \usepackage{cite}
\fi
\usepackage[table,xcdraw]{xcolor}
\usepackage{booktabs}
\usepackage{float}
\usepackage{amssymb}

\ifCLASSINFOpdf
\else
\fi
\usepackage{amsmath}

\usepackage{mathtools}
\usepackage{diagbox}
\usepackage{stfloats}

\DeclarePairedDelimiter\floor{\lfloor}{\rfloor}

\makeatletter
\newif\if@restonecol
\makeatother

\usepackage[linesnumbered,ruled,vlined]{algorithm2e}
\usepackage{algpseudocode}
\usepackage{amsmath}
\begin{document}
%
\title{FLFE: A Communication-Efficient and Privacy-Preserving Federated Feature Engineering Framework}
%
%
%
%

\author{Pei~Fang, Zhendong~Cai, Hui~Chen
        and~QingJiang~Shi
\IEEEcompsocitemizethanks{\IEEEcompsocthanksitem P. Fang, Z. Cai, and Q. Shi are all 
with the School of Software Engineering at Tongji University, Shanghai 201804, China.
Q. Shi is also with the Shenzhen Research Institution of Big Data, Shenzhen 518172, China. 
Email: shiqj@tongji.edu.cn\IEEEcompsocthanksitem H. Chen is 
with the Shenzhen Research Institution of Big Data, Shenzhen 518172, China.
\protect}\\
\thanks{Manuscript received April 19, 2005; revised August 26, 2015.}}

%
%

\markboth{Journal of \LaTeX\ Class Files,~Vol.~14, No.~8, August~2015}%
{Shell \MakeLowercase{\textit{et al.}}: Bare Demo of IEEEtran.cls for Computer Society Journals}
%



\IEEEtitleabstractindextext{%
\begin{abstract}
  Feature engineering is the process of using domain knowledge to extract features 
  from raw data via data mining techniques and is a key step to improve the performance 
  of machine learning algorithms. In the multi-party feature engineering scenario 
  (features are stored in many different IoT devices), 
  direct and unlimited multivariate feature transformations will quickly exhaust memory, power, 
  and bandwidth of devices, not to mention the security of information threatened. Given this, 
  we present a framework called FLFE to conduct privacy-preserving 
  and communication-preserving multi-party feature transformations. 
  The framework pre-learns the pattern of the feature to directly judge the usefulness of the transformation on a feature. 
  Explored the new useful feature, the framework forsakes the encryption-based algorithm for the well-designed feature exchange mechanism, 
  which largely decreases the communication overhead under the premise of confidentiality. We made experiments on datasets of both open-sourced and real-world thus validating the comparable effectiveness of FLFE to evaluation-based approaches, along with the far more superior eﬀicacy.

\end{abstract}

}

\maketitle

\IEEEdisplaynontitleabstractindextext

%
\IEEEpeerreviewmaketitle

\IEEEraisesectionheading{\section{Introduction}\label{sec:introduction}}

\begin{figure}[htbp]
  \centering
  \includegraphics[scale=0.16]{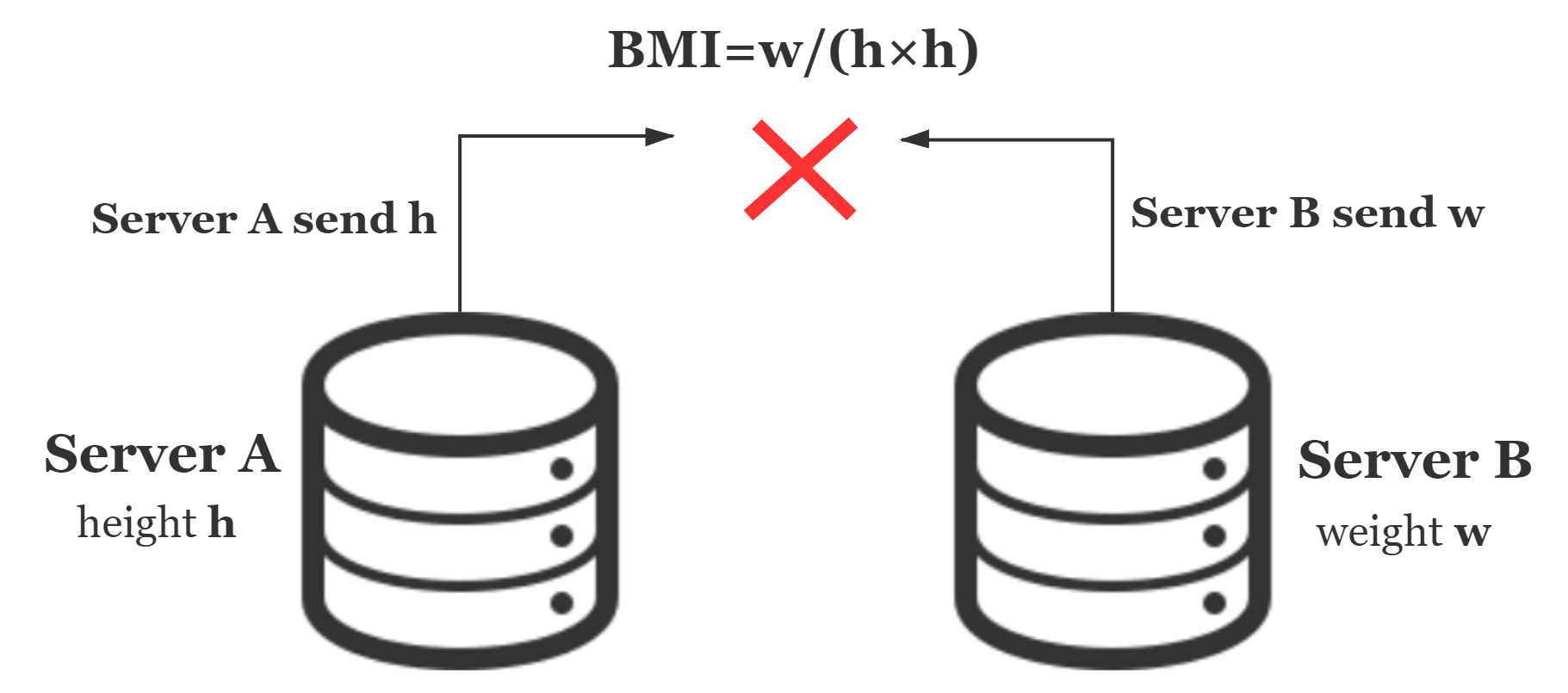}
  \caption{Direct exchange of users' data is not allowed. A straightforward solution is to encrypt $h$ and $w$, then caluclate $BMI$ with the ecrypted data.}
  \label{lawrestriction}
\end{figure}

\begin{figure}[htb]
    \centering
    \includegraphics[scale=0.28]{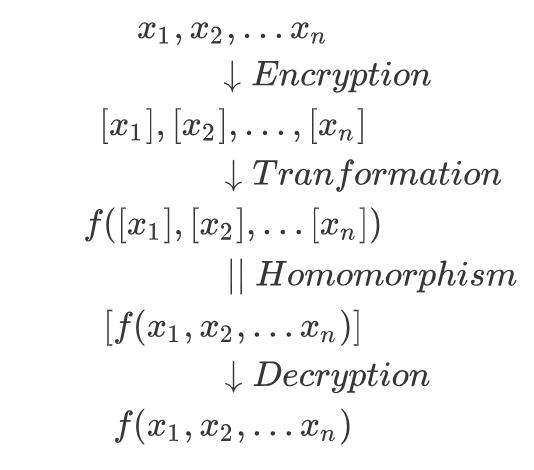}
    \caption{The application of HE in federated learning.}
    \label{fig-HE-FL}
  \end{figure}

%
%
%
%
\IEEEPARstart{F}{eature} Engineering is a key step in data preparation for machine learning. 
A feature engineering process usually includes applying transformation such as arithmetic and aggregation 
to original features. Appropriate transformations help scale the feature and significantly 
improve the performance of learning models. In traditional model-evaluation-based feature engineering, data scientists first select appropriate feature combinations to generate new features through domain knowledge,
and then conduct a specified machine learning model on datasets with and without the generated features to judge whether the generated features improve the model performance.\\
\indent In real-world applications, we often encounter multi-party feature engineering problem, where features come from different institutions and organizations. 
To preserve privacy, direct feature exchange for multi-party transformation is not allowed as it leaks users' privacy. In an example illustrated by Fig.~\ref{lawrestriction}, the Body Mass Index  ($BMI$) is to be calculated form height $h$ and weight $w$ by $BMI=w/h^2$, while $h$ and $w$ are separately collected by two sensors $A$ and $B$. However, under the law restriction, sensors $A$ and $B$ are not allowed to exchange their data without privacy guarantee. Besides privacy, communication overhead is also a big challenge. Frequent trials to generate new features, if few of the new features are judged useful, lead to vain communication with a huge overhead. \\
\indent To the best of our knowledge, there is no work in the literature that studies the secure multi-party feature engineering problem.
In this paper, we present a formulation of this problem and borrow the idea of federated learning to address this problem.\\
\indent Federated Learning incorporates concepts, methods, and applications on leveraging data from different parties to train a joint model without data leakage.
During the training process, each participant preserves its data locally, but upload the updated weights or gradients instead. 
According to how data overlaps, federated learning can be categorized into horizontal federated learning, vertical federated learning and transfer federated learning~\cite{yang2019federated}.
Vertical federated learning is applicable to the situation where the features of participants overlap less and the sample IDs overlap more, 
while the horizontal federated learning is the opposite. If both features and samples overlap very little among participants, 
federated transfer learning can apply the models learned in the source domain to the target domain.
In this paper we focus on the vertical federated feature engineering problem, 
in which participants share a large number of user IDs, but have disjoint user characteristics.\\
\indent The primary challenge of multi-party feature engineering problem is how to preserve privacy. 
A straightforward and widely used method to tackle this challenge is to apply encryption methods such as homomorphic encryption (HE) and differential privacy (DP).
HE encrypts the original text, and then performs various operations on the ciphertext to finally obtain the resulting ciphertext, 
which is expressed formally in Fig.~\ref{fig-HE-FL}.
The privacy preservation performance of HE depends on the length of the secret key, but longer secret key results in not only better performance but also higher computational complexity. In terms of application, Gentry~\cite{gentry2009fully} proposed the first available fully homomorphic encryption algorithm by introducing bootstrapping and squashing.
The disadvantages of Gentry's HE scheme are that Bootstrapping and Squashing are extremely computationally intensive, and that as the calculation proceeds the size of the ciphertext continues to expand.
Chai \emph{et al.}~\cite{chai2019secure} achieved a secure matrix factorization in recommendation systems through
homomorphic encryption on gradient information.
Different from HE, DP~\cite{10.1007/11761679_29} adds noise to the data to ensure minimal data leakage.
When the amount of data is small, the noise makes the final result far deviate from the accurate value. 
Also, a noiseless DP method was proposed in~\cite{bhaskar2011noiseless}, but it usually requires the data to follow some certain distribution,
which is difficult to satisfy in reality.\\
\indent Other than the privacy issue, communication overhead of multi-party feature engineering is also a big challenge. 
Most of the existing feature engineering approaches make all efforts to improve the data quality and enhance the performance, but rarely consider the possible mounting communication overhead in multi-party feature engineering scenario.
These methods usually adopt guided
search in feature space using heuristic feature quality measures~\cite{fan2010generalized,markovitch2002feature}, or 
perform greedy feature construction and selection based on model evaluation~\cite{dor2012strengthening,7837936}. 
The exhaustive and inefficient search renders the multi-party feature engineering costly in various aspects such as computing power, bandwidth and memory.
For example, in the context of the Internet of Things (IoTs), 
the edge devices always store and transmit data that expands with increase of users, 
and the above feature engineering approaches will lead to a high energy consumption and place a heavy burden on the devices, especailly those with small memory and low bandwidth.
To this end, many works focus on lossless compression through a process of encoding and then restoration~\cite{7737055,8249368}.
However these lossless methods either do not have a straightforward mapping to federated learning, or have a training process too complicated to adopt in practice.
Motivated by the limited resources of current devices, 
some recent works explored compression on different objects including gradients~\cite{konevcny2016federated,suresh2017distributed}, model broadcast~\cite{khaled2019gradient}, or local computation~\cite{caldas2018expanding}.
However usually the compression guarantees are instance
specific and depend on the entropy of the underlying distribution~\cite{cover1999elements}. In other words, if the data is
easily compressible, then they will provably be compressed heavily, and a statistical analysis on the correlation between compression standard and training effect is required.
Notably, recent works show that learning a
compression scheme in a data-dependent fashion also leads to a significantly better compression ratio~\cite{7149287,wu2017multiscale}.
However, it remains uncertain whether communication cost can be further reduced, and whether these methods or their
combinations can achieve the optimal trade-offs between communication and accuracy of feature transmission. \\
\indent To address the above two challenges, namely privacy and communication, we propose the Federated Learning Feature Engineering (FLFE), a framework to perform automated 
privacy-preserving and communication-efficient feature engineering. The idea behind FLFE is based on learning from 
labelled features besides the target dataset. At the core of the framework is a set of Multi-Layer Perceptron (MLP) classifiers, each corresponding to a specific transformation. 
Each classifier takes in Quantile Sketch Array (QSA)~\cite{Wang_quantilesover}, a fixed-sized array formed from feature(s), and judges whether the corresponding transformation will generate a useful new feature.
This procedure causes a problem in the federated setting: for a binary transformation, the two features to form a QSA may come from different participants. To this end, we present a communication mechanism among participants to keep the feature exchange confidential without the conventional encryption algorithms.
The mechanism, working well with the inherent characteristics 
of QSA, efficiently prevent one participant's features from leaking to the others. The main contributions of this paper are threefold.
\begin{enumerate}
  \item We use the QSA~\cite{Wang_quantilesover} as representation of feature. 
  QSA significantly reduces the communication overhead during multi-party feature transformations.
  In practice, the IoT device owners can easily 
  adjust the length of the QSA according to the performance and operating 
  conditions of their devices.
  \item We design a unique set of feature exchange mechanism.
  No homomorphic encryption or differential privacy is required.
  It preserves privacy with very few additional operations, and is efficient in computation and communication.
  \item We propose a framework which automatically performs feature engineering 
  without model evaluation. Since less time is spent on judging
  the usefulness of transformations, even brutally traversing all 
  possible combinations of features becomes feasible.
\end{enumerate}

\indent The remainder of this paper is organized as follows. Section 2 presents insight to the  privacy-preserving approaches in Federated Learning and automated feature engineering.
In Section 3, we define the multi-party feature engineering problem and give some necessary notations. 
In Section 4, we introduce some necessary components of FLFE, and then elaborate on the workflow of FLFE.
In Section 5, we conduct experiments on both open-sourced and real-world dataset. 
Finally we conclude the paper in Section 6.

\section{Related Work}\label{sec:relatedwork}
In this section, we highlight the stringent privacy requirements posed by federated learning. 
Also, we talk about existing techniques for feature engineering and possibility to utilize them to enhance the privacy and communication effectiveness of multi-party feature engineering. 
\subsection{Federated Learning}
Federated learning (FL) ~\cite{kairouz2019advances} is a scenario where multiple clients collaboratively train a machine learning (ML) model with the help of a central server.
Each client transfers local updates to the server for immediate aggregation, 
without having its raw data opened to other clients.
McMahan~\cite{DBLP:journals/corr/McMahanMRA16} first put forward the federated learning of deep networks based on iterative model averaging.
Based on the split of dataset, 
federated learning can be categorized into vertical FL, horizontal FL, and transfer FL.
In horizontal FL, 
the clients have different groups of data points and their features overlaps. 
In contrast, the clients share a joint group of data points with different features when it comes to the vertical one, 
of which the features become the input of deep feature transformations.
In this paper, we focus on the multi-party feature engineering problem in vertical FL.

\subsection{Privacy and Communication, challenges in federated learning}
 The main challenge is to balance the efficiency and confidentiality of communications.
Most methods in federated learning encrypt data to preserve privacy,
but complete encryption has a high demand for computing and communication resource.
Here, we first briefly introduce the methods for privacy preservation commonly used in federated learning.\\
\indent \textbf{Secure Multi-Party Computation (MPC)}~\cite{4568207} enables multiple participants to collaboratively compute an agreed-upon 
function with private data in a way that each party only knows its input and output (zero knowledge).
The participants agree on a function to compute, and then can use an MPC protocol
to jointly compute the output of that function on their secret inputs without
revealing them. Secure multi-party computing can provide strong confidentiality on the premise of zero knowledge.
However, for some scenarios, complex communication protocols with a significant amount of communication are required~\cite{mohassel2018aby3,mohassel2017secureml}, which renders
MPC inapplicable in practical settings.\\
\indent \textbf{Differential Privacy(DP)} is another popular tool combined with model averaging and SGD to facilitate secure FL~\cite{pathak2010multiparty}. 
Unlike MPC, DP ensures privacy of each individual sample in a dataset by adding noises from a specific distribution.
Although DP can be efﬁciently implemented, 
it exposes plain gradients to the central server during aggregation, which is likely to be recovered.
Meanwhile, the loss precision leads to an increase of communication overhead and training time.
As a result, the availability of data is greatly discounted, making this technology far from the general application~\cite{dwork2008differential}.\\
\indent \textbf{Homomorphic Encryption (HE)} allows a certain set of computation operations (e.g., addition) to be performed directly on ciphertexts and then decrypt the results to get the true values. 
Though having many advantages, HE schemes also have some major limitations.
The first is the restricted information space. Almost all HE schemes use integers~\cite{brakerski2011fully,coron2011fully}, so we need to convert the data to integers before encryption. 
Another limitation is the size of ciphertext which greatly increased after encryption. 
The third limitation is that more operations in encryption leads to more noise, making it harder for dycrption.
The last and most important is the lack of support for division operations. 
To sum up, currenlt only a limited number of operations like additions and multiplications are allowed on encrypted data~\cite{brakerski2014leveled}, 
while complex functions such as activation functions in neural networks are still not compatible with HE schemes.\\
\indent The above privacy-preserving approaches sacrifice communication efficiency for privacy.
In face of the mounting communication overhead, recent works on federated learning focus on compression of transmitted data. They have developed a series of technologies such as subsampling, problistic quantization and random mask~\cite{konevcny2016federated,liu2019enhancing}.
In fact, all of these technologies are looking for a universal solution for privacy preservation in federated learning.
However, for some special scenarios like multi-party feature engineering, 
if the original data has been converted into features that can be learned by the model before training, 
it is likely to protect privacy without encryption.
Next we list some recent feature engineering methods and analyze their applicability in federated learning.

\subsection{Feature Engineering}
Generalized and Heuristic-free Feature Construction~\cite{fan2010generalized}
avoids exhaustive enumeration of feature space by the divide-and-conquer strategy and weighting-rules-based search.
Markovitch \emph{et al.}~\cite{markovitch2002feature} found that the supplied set of attributes is not sufficient for creating an accurate, 
succinct and comprehensible representation of the target concept in classification task.
In view of this, they utilized heuristically beam search and predefined a set of feature constructors to constantly transform the original features. 
Admittedly, the above automated feature engineering approaches do not require domain knowledge, and brutally perform more feature transformations in than manual ones in the same time.
This characteristic renders them unsuitable for applications in the real world.
An automated feature engineering approach in a mullti-party setting should possess wisdom, that is, the ability to generate useful features with as few trials as possible.  
To improve the calculation efficiency, the more recent ExploreKit~\cite{7837936} performs automated feature engineering by combining information in
original features. Meanwhile, it restricted the exponential
growth of the feature space by learning both the enrtopy of new feature and the statistical tests on parent features.
This meta-learning-based approach inspired the optimization of feature information presentation.
Nargesian Konečný~\cite{nargesian2017learning} first proposed the QSA as input of meta-feature learners (models),
which shows an overwhelming advantage in representation capability over manual selection of features and two-layer auto-encoder.\\
\indent In summary, a viable multi-party feature engineering framework should work automatically
so that it can support more transformations in the same time.
Since the effect of feature engineering depends on the distribution of data, 
the framework should also have access to the true distribution as much as possible
under the guarantee of privacy with relatively lower communication overhead.
Admittedly, some approaches in federated learning do preserve privacy, but they either require high communication overhead (MPC, HE) or
sacrifce the precision of data transmission (DP). Motivated by the requirements for the above properties, we design FLFE, a federated feature learning framework that is automated, privacy-preserving and communication-efficient. The core idea of FLFE is the combination of automated feature engineering methods and a feature exchange mechanism, to which the section \ref{sec:sructurelfle} is devoted. 

\section{Multi-party Feature Engineering Problem}\label{sec:problemformulation}
Consider a set with $n$ devices $T \triangleq \{T_1, T_2, ... ,T_n\}$. 
Each device $T_k$ stores a dataset $D_k$, 
and the feature space of $D_k$ is defined as $F_k=\{f_{k1},f_{k2},...,f_{km_k}\}$, 
where $m_k$ is the number of features in $D_k$. 
We apply multi-party feature engineering among $\{T_1, T_2, ... ,T_n\}$.
According to the the application conditions,
multi-party feature engineering scenarios are divided into three levels:
\begin{enumerate}
  \item there is no requirement on confidentiality or communication efficiency;
  \item there is a requirement on confidentiality but no requirement on communication efficiency;
  \item there is a requirement on both confidentiality and communication efficiency.
\end{enumerate}
Among the three levels, the final one is the most demanding.
Thus, if we can solve the multi-party feature engineering problem in the final case, 
then there are nothing difficult in the others.
\subsection{No Requirement on Confidentiality or Communication}
\label{sec:3-1}
Suppose there is no requirement on confidentiality and communication, 
a simple idea is to transmit the data scattered across the devices in plain context, 
so that we obtain $D=D_1 \bigcup D_2 ... \bigcup D_n.$
Data scientists apply appropriate transformations on features based on domain knowledge, 
and then use model evaluation methods such as random forest, logistic regression, etc. to verify 
the effect of the transformation. Except for the procedure of feature transmission, 
multi-party feature engineering in this scenario is nothing different from the traditional feature engineering.
\subsection{A Requirement on Confidentiality, No Requirement on Communication}
\label{sec:3-2}
When there should be privacy preservation during transmission
but no strict restrictions on communication overhead, the problem is amenable to encryption.
Typically an encryption-based procedure conduct multi-party feature transformations as follows:
Participants encrypt features with Homomorphic Encryption and transmit them to a server;
The server performs transformations and decrypts the result, and the decrypted result
is equivalent to the result of direct transformation without encryption;
The server finally decide on whether to retain or abandon these features. Notice that a disadvantage of this procedure is that in practice, because of encryption, the transformations are limited to combinations of additions and subtractions or that of multiplications and divisions. 
\subsection{A Requirement on Both Confidentiality and Communication}
\label{sec:3-3}
Since the ciphertext generated by homomorphic encryption is a long byte string (usually 512 bits for one integer), transmission of the ciphertext results in a huge waste of resources, 
both in computation and communication.
At the same time, generation of secret key during encryption is extremely time-consuming.\\
\indent To further illustrate how expensive the feature engineering approaches in Section \ref{sec:3-1} and \ref{sec:3-2} are, 
we consider $n$ devices each with $m_k$ features, where $1\leq k\leq n$. 
We need to apply a binary transformation of features from two different devices, so there are $\frac{1}{2} \times ((\sum^n_{d=1} m_d)^2-\sum^n_{d=1} m_d^{2})$ possible choices of the two features.
Suppose $b$ binary transformations are available for selection,
there are $\frac{1}{2}b \times ((\sum^n_{d=1} m_d)^2-\sum^n_{d=1} m_d^{2})$ possible new features. 
With $b$ fixed, the amount of new features and their combinations to explore grows rapidly. 
Hence the coventional routine of encryption, transmission, model evaluation, and then enumeration of all the transformation combinations, is not practical.
A scalable method must avoid 
this computational bottleneck. \\
\indent We design FLFE in the situation where participants 
share the same set of samples but with different features, which is consistent with the definition of vertical federated learning introduced in Section 1.
We emphasize that FLFE aims to solve the multi-party feature engineering problems of the most difficult level.

\section{The Structure of FLFE}\label{sec:sructurelfle} 
\begin{figure}[htb]
  \centering
  \includegraphics[scale=0.14]{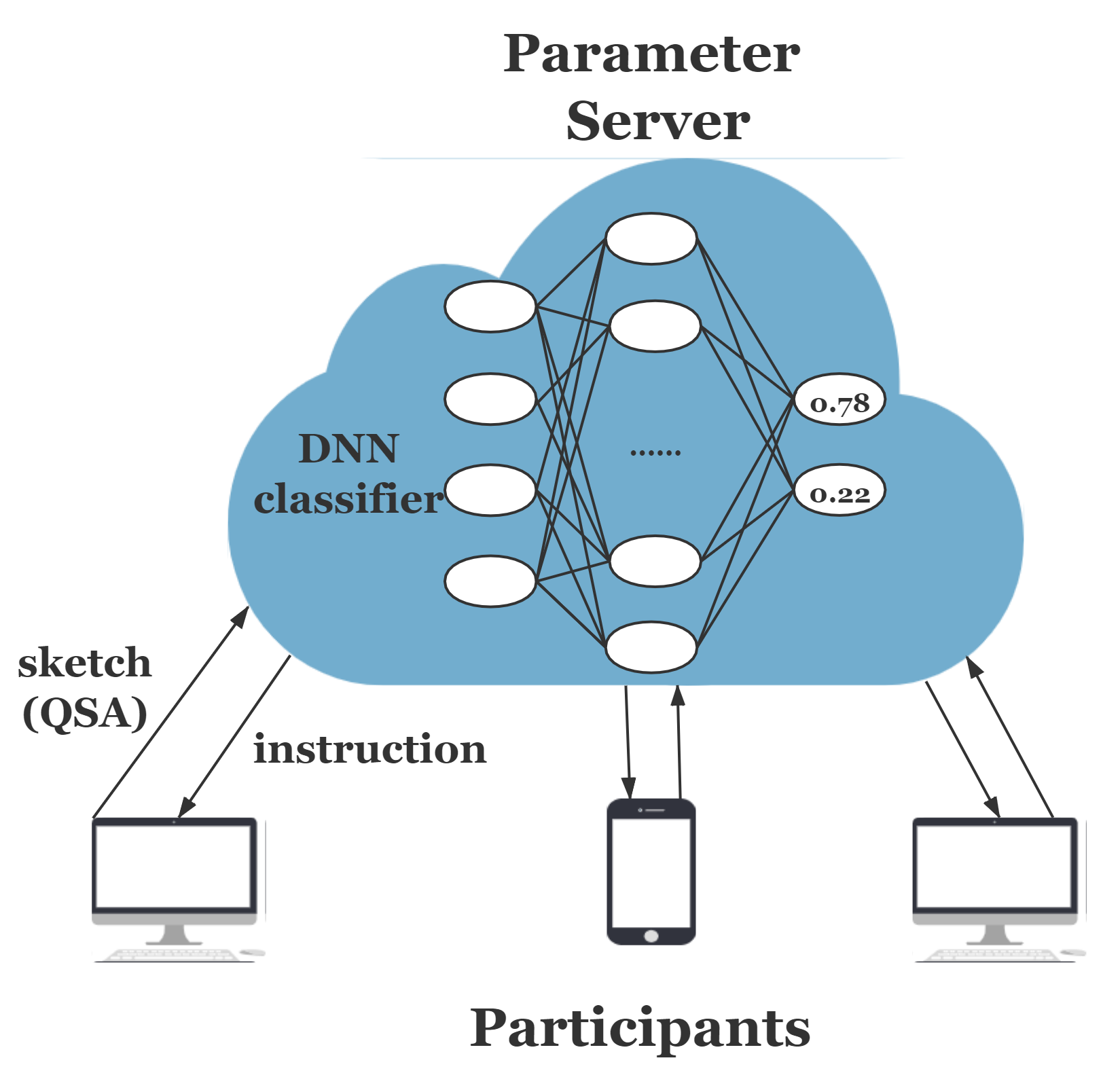}
  \caption{The overview of FLFE structures.
  Participants upload sketchs (QSA, which will be elaborated in Section 4.1) to the server, and the server in turn issues instructions to coordinate the operation of multi-party feature engineering. 
  DNN Classifiers tell whether a feature is useful from the sketchs uploaded.
  Ideally, participants can be all kinds of communication devices that hold features}
  \label{structure}
\end{figure}

\indent In FLFE, our vision is to bring to light the integration of sketches and DNN to practical multi-party feature engineering problems
and show a viable path towards a federated feature engineering system.
As the Fig.~\ref{structure} shows, the center of FLFE is a parameter server, a platform laid a set of Deep Neural Networks,
with the participants in multi-party feature engineering scattered around.
Communications between the device and the server, and between the device and the device 
are both maintained under certain bandwidth limitations.
Besides the storage of DNN models, the parameter server sends instructions 
to coordinate feature exchange of devices and accept the sketches generated by participants.
Since the multi-party feature engineering system in Section 3.2 requires participated devices to support
complicated encryption-based privacy-preserving methods, here in FLFE, we assume that
all the devices are at least able to perform operations with $O(n)$ time complexity.\\
\indent In a nutshell, FLFE generates QSAs from features, transmit them, and feed them into a trained DNN; Then if the DNN outputs a result larger than a threshold, which means the transformation (corresponding to the DNN) works well on the features, the original features will be securely transmitted and transformed into a new feature. In this section, we first discuss about some essential components of FLFE, and elaborate on FLFE's workflow in the last subsection. 
The components include the QSA (Section~\ref{feature-representation}), DNNs and their training (Section~\ref{modelserlectionandtraining}), transformation judgement and feature generation (Section~\ref{transformation-judgement}).
\subsection{Feature Representation}
\label{feature-representation}

In FLFE, we use the Quantile Data Sketch(QSA)~\cite{Wang_quantilesover}, instead of the original features, as input of DNN Classifiers for three reasons: (1) it does not conform to the principal of privacy preservation to transmit original features; (2) the DNN classifiers have a fixed dimension of input, but the datasets vary in number of data points; (3) Nargesian \emph{et al.} ~\cite{nargesian2017learning} found that, in terms of predictive performance, QSA is well above other representations such as Hand-crafted Meta-features, Stratiﬁed Sampling, and Meta-feature Learning.

\begin{figure}[hbtp]
  \centering
  \includegraphics[scale=0.1]{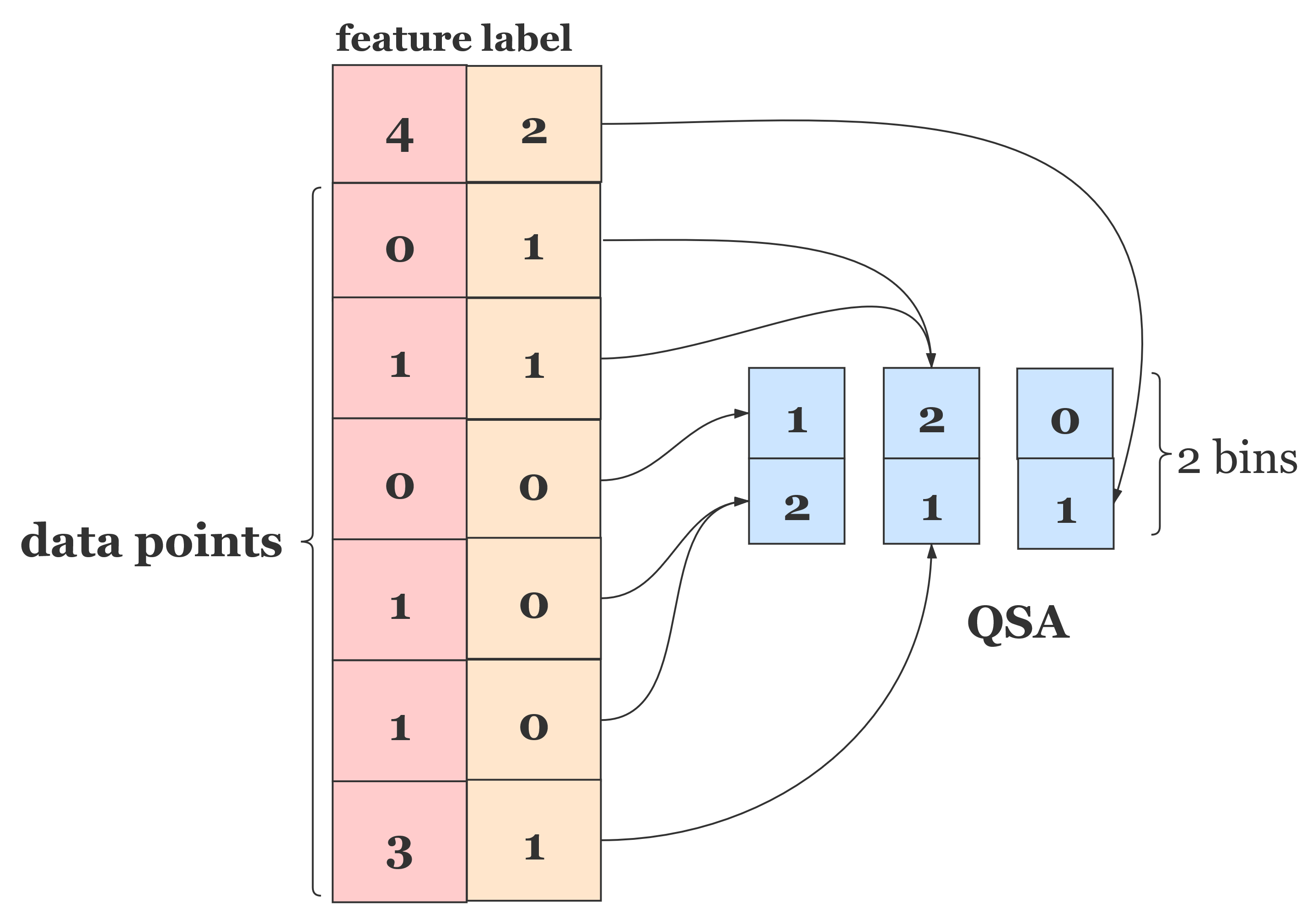}
  \caption{An example of the generation of QSA from a feature. The label is either $0$, $1$ or $2$,
  so the QSA has three columns. The data points labelled as $0$, $1$, and $2$ are correspond to the first, second, and third column seperately.
  In this example, the number of bins is set to 2 and the actual bin index for each data point is calculated in equation (1) and (2).
  Generally, QSA can reveal the correlation between the distribution of a feature and the labels.}
  \label{qsageneration}
\end{figure}

A feature (of size $num~data~points \times 1$) is transformed into a QSA (of size $m \times num~classes$) with the help of labels, where $m$ is a manually set parameter. A column of a QSA corresponds to the distribution of data points in a certain class with respect to this feature. To be specific, a feature $f$ is transformed into a QSA:
\begin{gather*}
    R_f = [S_f^1,S_f^2,...,S_f^n]\\
    \mbox{where $n$ is the number of classes,}\nonumber \\
    \mbox{and $m$ is the length of each column $S_f^k(k=1,\cdots,n)$.}\nonumber
\end{gather*}
To calculate $S_f^k$, suppose there are $t$ data points with label $k$, and they are $k_1,k_2,\cdots,k_t$, with features $f_k \triangleq \{f_{k_1},f_{k_2},...,f_{k_t}\}$.
We define $m$ disjointed bins with size $b_0,b_1,...,b_{m-1}$, which are initialized to $0$ and should finally satisfy $\sum_{k=0}^mb_k=t$. 
For $v\in\{1,2,\cdots,t\}$, we put data point $k_v$ in the $id$-th bin, as
  \begin{gather}
    \label{hahah}
    id = \floor*{\frac{f_{kv}-\min{f_k}}{\max{f_k}-\min{f_k}} \times m} \\
    b_{id}=b_{id} + 1 \\
    \mbox{where $\floor*{\cdot}$ indicates rounding down.}  \nonumber
\end{gather}
Notice that $k_v$, with $f_{k_v}=\max{f_k}$, cannot be assigned according to the above formula, so it is directly put in $b_{m-1}$. 
Repeat the above process for $k=1,\cdots,n$ to obtain $R_f$. An example of QSA generation is given in Fig.~\ref{qsageneration}.
\\
\begin{figure}[ht]
  \centering
  \includegraphics[scale=0.12]{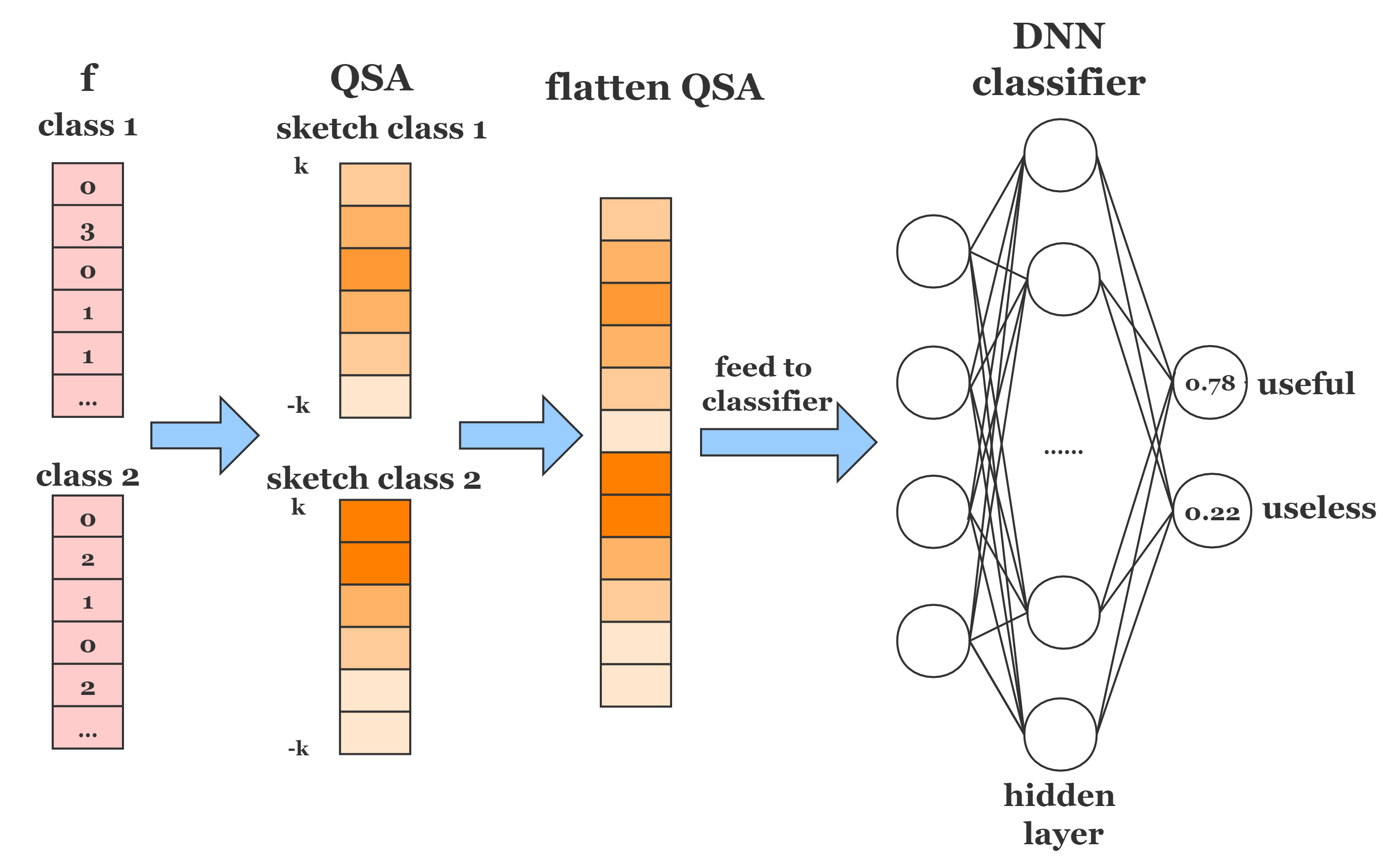}
  \caption{The process from sketching a feature to getting an output
  of DNN classifier. 
  a QSA is generated from a feature, 
  scaled to $[-k,k]$ separately for each class, flattened, and fed into DNN classifiers.
  The depth of color in QSA
  represents the number of data points put into the certain bin.
  The output has a confidence score measuring the probability of the transformation (corresponding to the classifier) being useful on the feature $f$. 
  In practice, we can adjust QSA parameters (number of bins and scaling range) based on the memory and bandwidth of devices.}
  \label{qsa_nn}
\end{figure}

  \indent To further illustrate how a QSA is utilized by DNN Classifiers, Fig.~\ref{qsa_nn} demonstrates the process from the generation of a QSA to the ouput of a classifier. 
  In order to fix the input size of DNN classifiers for different classification tasks, 
  we convert the multi-class problems into one-vs-all~\cite{rifkin2004defense} binary classiﬁcation problems. 
  A QSA manages to convert a feature with variable length (being equal to the number of the data points) to a fixed-length array acceptable for Neural Networks. 
  Since the true values of features are replaced by QSA, 
  the participants can securely transmit the QSAs without encryption. Besides,
  since a feature might be as long as tens of thousands, 
  the application of QSA will also reduce a large amount of communication overhead.
  
  \subsection{Model Selection and Training}
  \label{modelserlectionandtraining}
  On the selection of models, our insight is an integration of Deep Neural Network (DNN) and sketches.
  The superior predictive performance of DNN stems from extraction of high-level features 
  of data. However, 
  DNN is criticized for the lack of interpretability,  
  the ability to let a layperson to understand why a DNN delivers certain results.
  The sketch is mainly limited in application to the areas of network measurement and databases.
  We find that through integration, the defects of sketches and DNN
  interestingly reverse to optimize \textit{two sides
  of the same coin} (performance and privacy in multi-party feature engineering). 
  There are two reasons:
  (a) First, the sketch, usually a fixed-sized array, 
  attains high accuracy with succinct data structure, 
  and reach an adjustable trade-off between accuracy and memory. 
  This is important in attaining learning efficiency and prevent
  battery draining and overheating problems of mobile phones and other devices. 
  (b) Second, we notice that the cooperation of DNN and sketches
  brings about inherent but little explored privacy benefits.
  A sketch compresses data length but preserves the feature of data so that
  It is difficult for people to extract information from a sketch,
  while the trained DNN can easily accomplish these tasks.
  Since DNN is well-known for its `Black Box' characteristic, 
  it is hard to know the basis of DNN inference.\\
  \indent The training of a DNN classifier requires sufficient labelled QSAs as training samples. 
  We applied transformations only on numerical features in classiﬁcation datasets, but not on discrete ones, to generate training samples (QSAs without labels).
  For a transformation, the training samples are labelled as positive if the transformation improves the performance of a base model on the dataset.
  Specifically, we decide the standard of improvement by evaluating a base model $L$ (Random Forest, Logistic Regression, etc.) on the original feature and the constructed feature. 
  If the constructed feature leads to a performance improvement beyond a threshold, 
  the training sample will be labelled as positive. 
  Because these QSAs are all from open-sourced dataset, it's not necessary to keep them confidential, and the training of DNNs can be done in the parameter server.\\
  \indent As for the selection of DNN structure, we have tried many types of neural network structures, which vary in number of cores and depth of hidden layers.
  Our trials showed that even a simple Multi-Layer Perceptron (MLP) binary classiﬁer performs well on QSA prediction, and the training loss converges rapidly.
  This result again indicates QSA's strong ability to extract features and easy-to-learn data structure.
  In our experiment, we used MLPs with only one hidden layer.
  For a QSA representation $R_f = [S_f^1,S_f^2,...,S_f^n]$, the probability of being useful is computed as:
    \begin{gather*}
  \begin{array}{l}{\left[p_{useful}(f), p_{useless}(f)\right]=} \\ \sigma_{2}\left(\mathbf{b}^{(\mathbf{2})}+\mathbf{W}^{(\mathbf{2})}\left(\sigma_{1}\left(\mathbf{b}^{(\mathbf{1})}+\mathbf{W}^{(\mathbf{1})}R_f \right)\right)\right)\end{array}
  \end{gather*}
Here $\sigma_{2}$ and $\sigma_{1}$ are respectively softmax and rectiﬁed linear unit (ReLU) functions~\cite{nair2010rectified};
$\mathbf{W}^{(\mathbf{1})}$ and $\mathbf{W}^{(\mathbf{2})}$ are weight matrices;
$\mathbf{b}^{(\mathbf{1})}$ and $\mathbf{b}^{(\mathbf{2})}$ are bias.
In the training procedure, we adopt Adam Optimizer~\cite{kingma2014adam} and introduce both regulation and drop-out~\cite{hinton2012improving} with $0.5$ drop-out rate to prevent over-fitting.\\

\subsection{Transformation Judgement} 
\label{transformation-judgement}
In this section, we elaborate on how we judge transformations and generate new features in FLFE. Only unary and binary transformations are considered as $r$-ray transformations ($r>2$) can be composed by unary and binary transformations. 
There are multiple $r$-ary transformations, each corresponding to a trained DNN. For $r$ selected features $[f_1,f_2,...,f_r]$ and each $r$-ary transformation, we input the QSA generated from $[f_1,f_2,...,f_r]$ into the corresponding DNN and judge whether the transformation will produce a useful new feature.
As is desecribed in Section $4.2$, 
each DNN classiﬁer outputs a real-valued conﬁdence score, namely the probability of a transformation being useful.  
If the conﬁdence score is above a given threshold, FLFE recommends the corresponding transformation to be applied on features $[f_1,f_2,...,f_r]$.
In consideration of the prediction error and risk of dimension explosion,
we only recommend transformations which are very likely to be useful, and those ambiguously judged as useful (with confidence score around 0.5) will be abandoned.

\subsection{FLFE Workflow}
Based on the components introduced above, we now finally move on the workflow of FLFE.\\
\indent Judgment of validness of feature transformation 
requires computation of the QSA. It is only when a feature's QSA is judged as useful will FLFE actually generate features with the corresponding transformation. The enhanced privacy-preserving ability in the feature generation process is achieved mainly by the Parameter Server, which serves as an intermediary for information transmission.
Parameter Server~\cite{ho2013more} is a typical element in distributed machine learning, which 
is responsible for data storage from distributed working nodes, and allocation of computing resources through a central scheduling node.
To enhance confidentiality during feature generation and transmission, we introduce a similar server-client structure into FLFE: the parameter server is set up for scheduling feature engineering and transformation, and the clients 
are the devices scattered around the server, as feature-holders.
Different from distributed machine learning, the work of parameter server includes
deciding on transformation types, designating the participators, coordinating information exchange, and storing the 
generated features. Generally, all participants in FLFE has full autonomy
for the local data, and can decide when and how to join the federated feature engineering.\\  
\indent In FLFE, we define a loop as an attempt to validate the usefulness of a transformation on two features from different participants.
The final output, a confidence score, measures whether the the new generated feature is useful or not. 
Parties inolved in a loop include a Parameter Server and at least two participants (the feature holders).
The main procedure of FLFE with a parameter server and the two participants is shown in Algorithm.~\ref{FLFE_procedure}, with more details elaborated in Algorithm.~\ref{Judge_a_QSA} and Algorithm.~\ref{Generate_a_New_Feature}.
\begin{algorithm}[htbp]
  \caption{The procedure of FLFE}
  \label{FLFE_procedure}
  \LinesNumbered
  \KwIn{parameter server $Sr$ that holds new features and coordinate the feature exchange,
  a set of participants $dcs=\{dc_1, dc_2,..., dc_n \}$}

  $S_r$ gets the information of $\{dc_1, dc_2,..., dc_n \}$\;
  $S_r$ sets the $MaxLoop$\ and $\theta$\;

  \If {$Sr$ and $dcs$ have prepared}
  {
    \For{$loop=1;loop \le MaxLoop;loop++$}
    {
      $prob = Judge\_a\_QSA(S_r,dcs)$\;
      \If {$prob \geq \theta$}
      {
        $Generate\_a\_New\_Feature(S_r,dcs)$\;
      }
    }
  }
  \textbf{final} \;
  \end{algorithm}

\indent As Algorithm~\ref{FLFE_procedure} shows, the whole procedure of FLFE does not end until the number of completed loops reaches the $MaxLoop$. 
In order to maintain the stability and efficiency of FLFE, participants in poor conditions can notify the server and then exit the loop,
or the server selects the devices in good condition in priority during feature engineering.
Generally, the workflow of FLFE can be described as a `two-step' strategy.
It is only after the first step($Judge\_a\_QSA$) returns a confidence score
higher than the given threshold $\theta$, 
will a new feature genuinely be generated(the second step). Otherwise, Algorithm~\ref{FLFE_procedure} restarts as a new loop begins.\\
\indent Algorithm~\ref{Judge_a_QSA} depicts the process of judging a QSA under the coordination of the Parameter Server.
Algorithm~\ref{Generate_a_New_Feature} demonstrates the privacy-preserving feature exchange process.
It includes steps from generating a new feature among participants to sending it to the Parameter Server.
Fig.~\ref{flfe_diagram} gives an illustration of the departure and destination of each feature transmission in the two algorithms.  \\
\begin{figure}[htbp]
  \centering
  \includegraphics[scale=0.28]{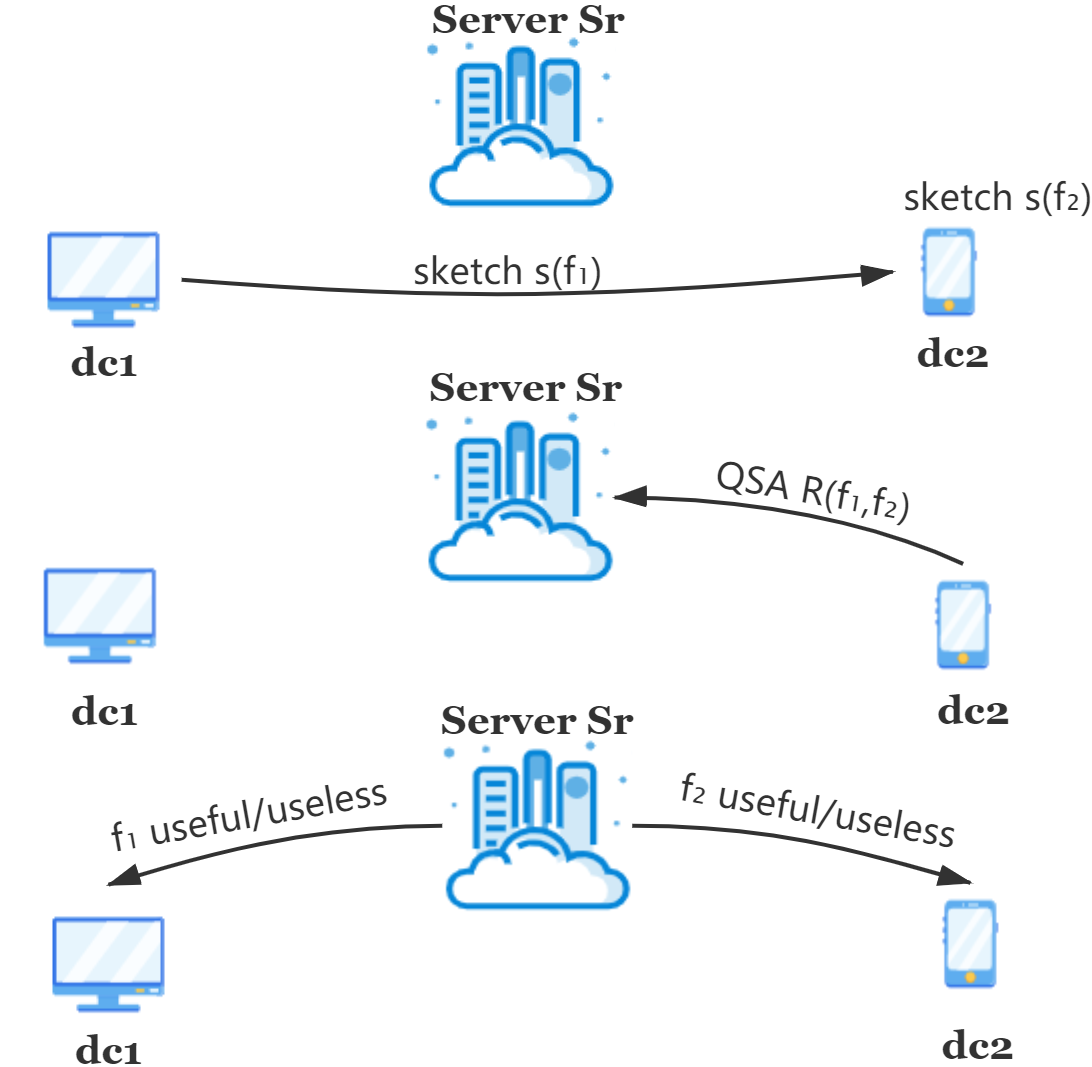}
  \caption{Illustration of Algorithm~\ref{Judge_a_QSA}, which judges the QSA of original features.
  Some details besides data transmission are elaborated:
  In the second picture, $dc_2$ first flatten the $s(f_1)$ and $s(f_2)$ to generate the $QSA\ R(f_1,f_2)$.
  In the third picture, the server $S_r$ tells whether a transformation generates a new feature.}
  \label{flfe_diagram}
\end{figure}
\begin{algorithm}[htbp]
  \caption{$Judge\_a\_QSA$}
  \label{Judge_a_QSA}
  \begin{algorithmic}[1]
    \Require
    parameter server $Sr$;
    a set of participants $dcs=\{dc_1, dc_2,..., dc_n \}$;
    \Ensure
      The probability $\theta$ of a feature being useful;
    \State Server $Sr$ decides on which two features, $f_1$ in deivce $dc_1$ and $f_2$ in device $dc_2$, are to participate in this loop;
    \label{code:fram:extract}
    \State Server $Sr$ selects a transformation $T$ (Sum, Multiplication, etc.) 
    and notify $dc_1$ and $dc_2$ the transformation type and $f_1$, $f_2$ separately;
    \label{code:fram:trainbase}
    \State Device $dc_1$ generates sketch $s(f_1)$ and send $s(f_1)$ to $dc_2$;
    \label{code:fram:treain}
    \State Device $dc_2$ generates sketch $s(f_2)$, obtain the QSA $R(f_1,f_2)$ and send $R(f_1,f_2)$ to $Sr$;
    \label{code:fram:add}
    \State Server $Sr$ feeds $R(f_1,f_2)$ into the DNN classifier corresponding to T, output the probability $\theta$ judging whether new feature $f_3=T(f_1,f_2)$ will be useful;
    \label{code:fram:classify}
    \State Server $Sr$ tells $dc_1$ and $dc_2$ about the judgement. If useful, both $dc_1$ and $dc_2$ prepare for the next step;
    \label{code:fram:select} \\
    \Return $\theta$;
  \end{algorithmic}
\end{algorithm}

\indent In Algorithm~\ref{Judge_a_QSA}, $dc_1$ first transmits the sketch of its feature $f_1$ and sends it to another device $dc_2$. Then $dc_2$ combines the received sketch with a sketch of its own feature $f_2$ to form a QSA.
The formed QSA is sent to the Parameter Server, which then use a DNN to judge whether the formed QSA indicates a useful new feature and return the confidence score.\\
\begin{figure}[htbp]
  \centering
  \includegraphics[scale=0.275]{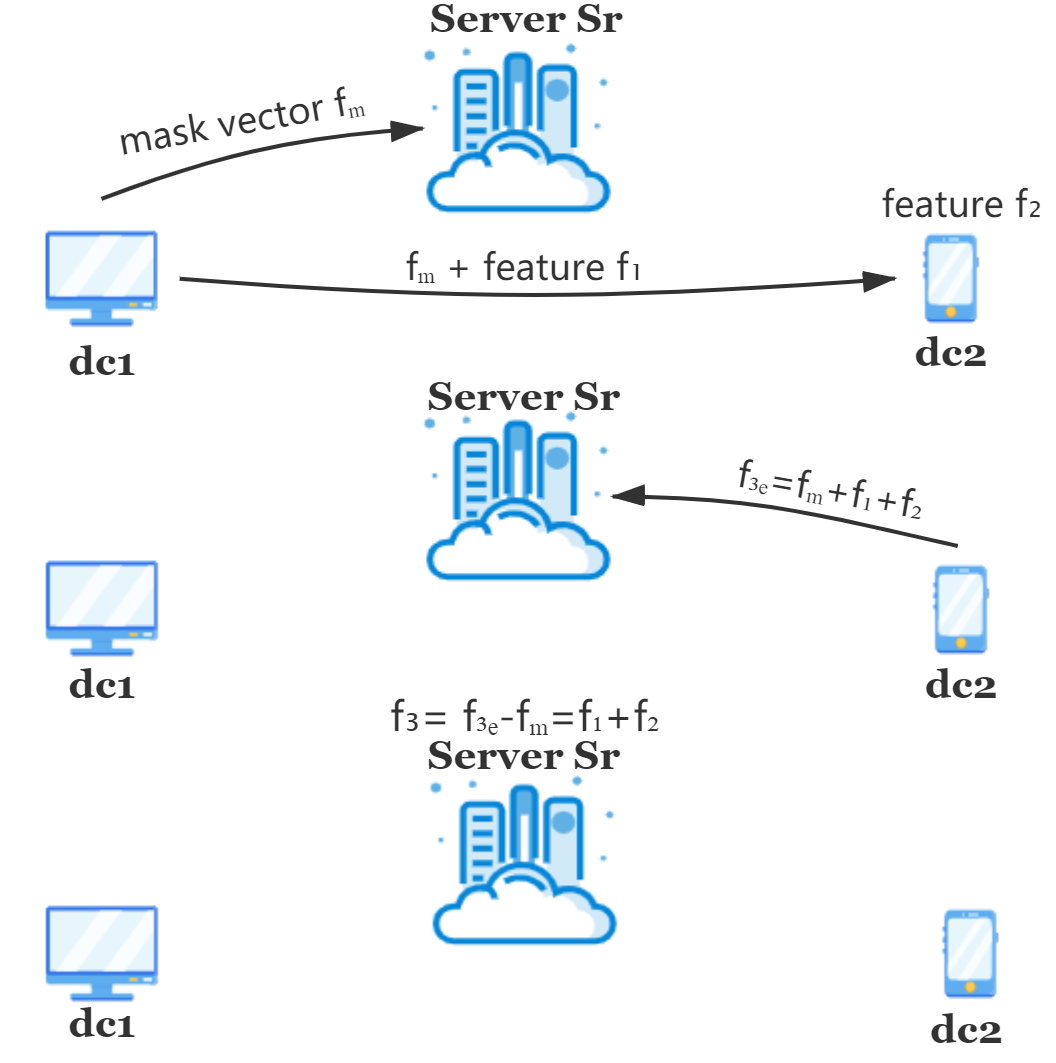}
  \caption{Illustration of Algorithm~\ref{Generate_a_New_Feature} (in this example, Addition is taken as the transformation $T$), which generates a new feature and stores it in the Parameter Server.
  Some details besides data transmission are elaborated:
  In the second picture, $dc_2$ first calculate the $f_{3e}=f_1+f_2+f_e$. The $f_1+f_m$ comes from $dc_1$ and $f_2$ is stored by $dc_2$ itself.
  In the third picture, the server $S_R$ restored the true value $f_3=f_{3e}-f{e}$. The $f_{3e}$ comes from $dc_2$ and $f_m$ is the mask vector in the first picture.
}
  \label{flfe_diagram}
\end{figure}
\begin{algorithm}[htbp]
  \caption{$Generate\_a\_New\_Feature$}
  \label{Generate_a_New_Feature}
  \begin{algorithmic}[1]
    \Require
    parameter server $Sr$;
    a set of participants $dcs=\{dc_1, dc_2,..., dc_n \}$;
    \State Device $dc_1$ encrypts $f_1$ with a random mask vector $f_m$, thus getting the encrypted feature $f_{1e}=encrypt(f_1, f_m)$;
    \label{code:fram:extract}
    \State Device $dc_1$ sends $f_{1e}$ to $dc_2$;
    \label{code:fram:trainbase}
    \State $dc_2$ calculates $f_{3e} = T(f_{1e},f_2)$;
    \label{code:fram:add}
    \State Device $dc_2$ sends $f_{3e}$ to $Sr$;
    \label{code:fram:classify}
    \State Server $Sr$ gets $f_m$ from $dc_1$;
    \label{code:fram:select} 
    \State $Sr$ gets the final new feature $f_3$ by decrypting $f_{3e}$ with $fm$, which is expressed as $f_3=decrypt(f_{3e},f_m)$;
  \end{algorithmic}
\end{algorithm}
\indent Algorithm~\ref{Generate_a_New_Feature} proceeds only if the confidence score returned by Algorithm~\ref{Judge_a_QSA} is larger than the preset threshold.
To conceal true values of the feature in $dc_1$, we initialize a random mask vector $f_m$, which is a vector with the same length as $f_1$ and $f_2$.\\
\indent $f_3$ in Algorithm.~\ref{Generate_a_New_Feature} should ideally equal to $T(f_1,f_2)$,
so the $encrypt$ and $decrypt$ operations should satisfy:
\begin{align}
  T(f_1,f_2)&=decrypt(f_{3m},f_m)\\
      &= decrypt(T(f_{1m},f_2),f_m)\\
      &= decrypt(T(encrypt(f_1,f_m),f_2),f_m), \label{last-Tf1f2}
\end{align}
where equation (\ref{last-Tf1f2}) corresponds to the whole procedure demonstrated in Algorithm~\ref{FLFE_procedure}. The $encrypt$ and $decrypt$ operations vary from transformation to transformation.
If binary operations are sum, subtraction, multiplication or division, and both the $encrypt$ and $decrypt$ operations satisfy
the following equations, the target equation (\ref{last-Tf1f2}) can be met.
\begin{align}
  T &= encrypt; \label{encrypt}\\
  f_o &= decrypt(encrypt(f_o,f_m),f_m).\label{decrypt}
\end{align}
\begin{align}
\intertext{If T is addition, subtraction, multiplication, or division,}
  &T(T(f_1,f_2),f_m) = T(T(f_1,f_m),f_2).\\
\intertext{From (\ref{encrypt}):}
  \Rightarrow &encrypt(T(f_1,f_2),f_m) = T(encrypt(f_1,f_m),f_2).\\
\intertext{From (\ref{decrypt}):}
  \Rightarrow &T(f_1,f_2) = decrypt(T(encrypt(f_1,f_m),f_2),f_m),
\end{align}
which is exactly equation (\ref{last-Tf1f2}). Notice that unlike Homomorphic Encryption, where there is a strict requirement on the encryption, here encryption can be as simple as addition or multiplication, and the decryption is correspondingly subtraction or division.\\
\indent In summary, FLFE adopts a `two-step' strategy
to address the multi-party feature engineering problems. The first step only requires the transmission of QSA instead of the true feature value.
In face of limitless data points and feature combinations, most of which are useless, such initiatives avoid a large amount of fruitless server-participant and participant-participant communication cost. Furthermore, the Parameter Server organizes the 
FLFE workflow but not decides on how to encrypt and decrypt features, or the initialization of mask vector $f_m$.
In FLFE, each participant keeps its local feature secret to others, and the Parameter Server only stores the generated new features.
That is to say, FLFE not only reduces the communication overhead but also provides a strong privacy guarantee.\\
\indent Compared with traditional feature engineering approaches, we attribute the relatively lower communication overhead to the size of QSA.
Meanwhile, the privacy guarantee of FLFE is result of the joint effect of feature representation, data transmission design, and DNN's lack of interpretability.

\section{Experiment}
\begin{figure}[hbtp]
  \centering
  \includegraphics[scale=0.124]{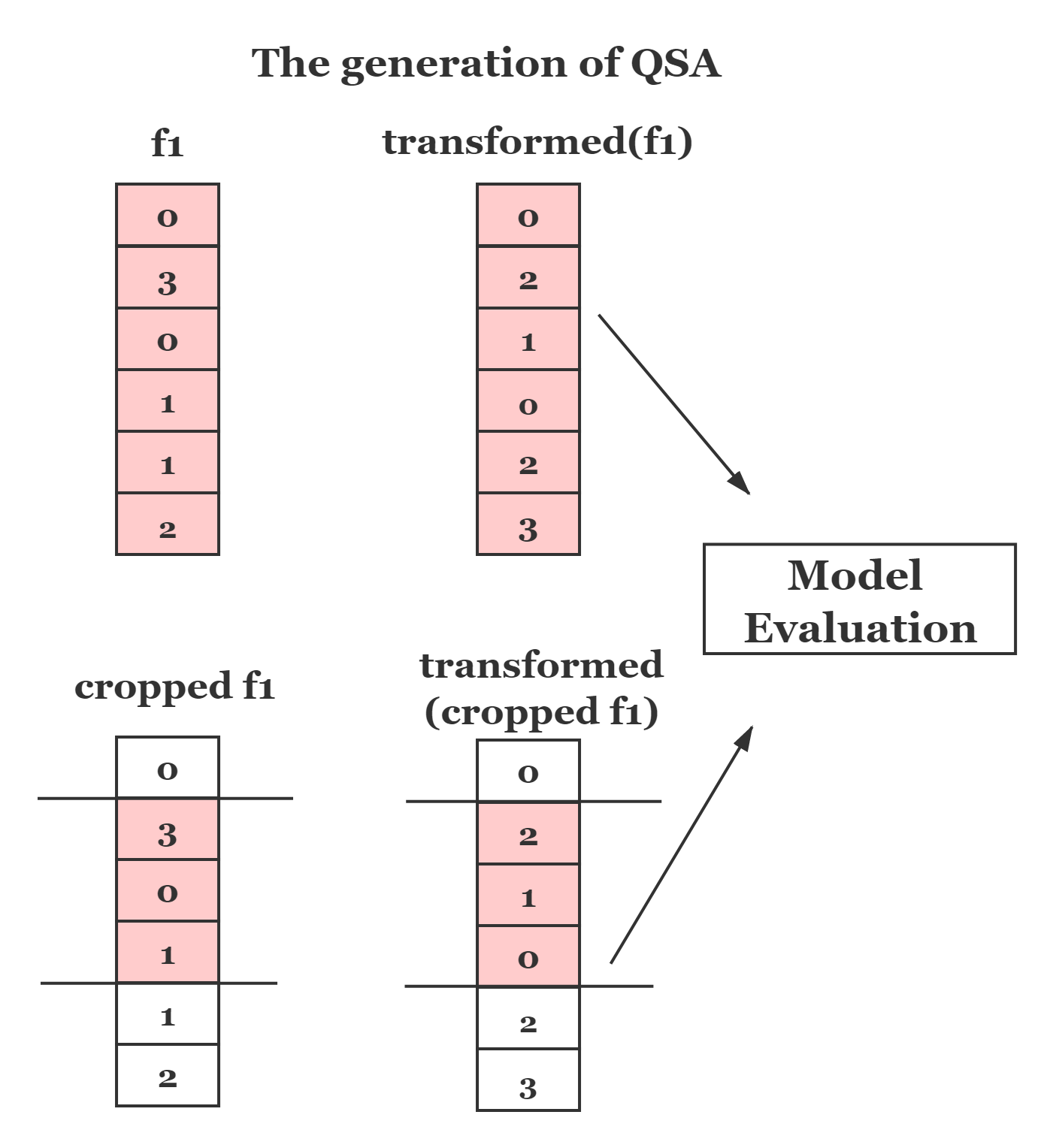}
  \caption{We cropped a feature at a random rate multiple times to generate more training samples (QSAs).}
  \label{crop}
\end{figure}

\indent We evaluate the performance of FLFE in two experiments: 
(1) efficacy on classification tasks compared with (1.a) instant model evaluation and (1.b) learning feature engineering (LFE) conducted separately on each device;
(2) efficiency in terms of computation time and communication overhead.
The former was done on open-sourced datasets, each manually spilted into three sub-datasets with same data points but different features, while the latter on a real-world dataset of the vertical federated learning setting.\\ 
\indent In the first experiment, we generate all feasible combinations of features for binary transformations in advance to eliminate the error caused
by random feature transformations. In other words, the three approaches always transform the same features and the difference of classification result come from the approaches themself.
The second experiment doesn't have such a setting, which means the generated new feature can be further transformed, 
so that the experiment can better present the actual efficiency in running time and communication overhead.\\
\indent We implemented FLFE transformation classiﬁers in Pytorch and simulate the transmission of QSA and true features in PySyft. 
Transformation and instant model evaluation were implemented with Scikit-learn. 
For FLFE, we considered the following 14 transformations: 
log, square-root (applied on the absolute of values), frequency (count of how often a value occurs), square, round, tanh, sigmoid, isotonic regression, zscore, normalization (scaling to [-1, 1]) for unary transformations,
sum, subtraction, multiplication and division for binary ones.  \\
\begin{figure}[htb]
  \centering
  \includegraphics[scale=0.152]{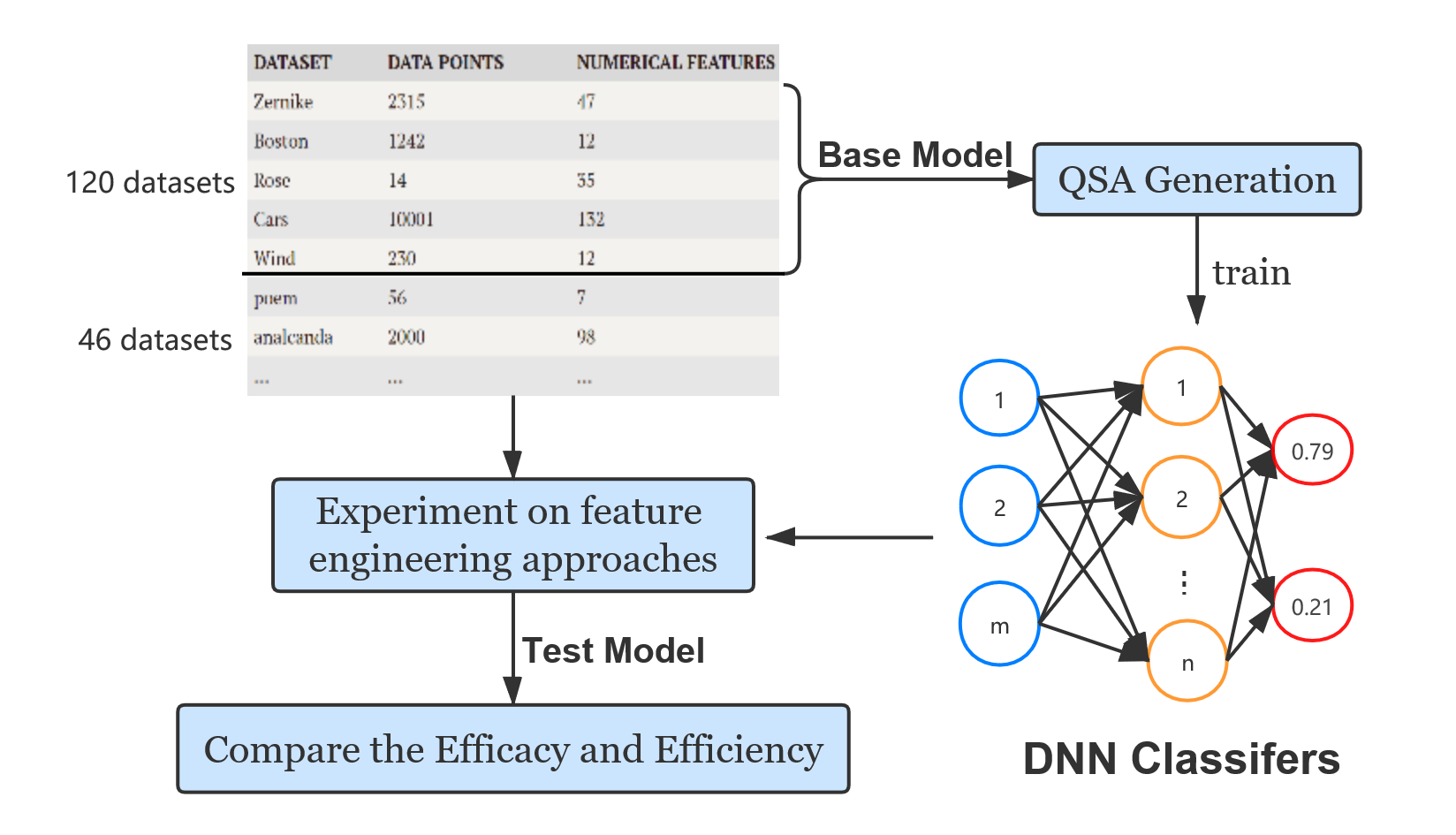}
  \caption{The procedure from data preparation to experiment conduction. We generate QSAs from 120 open-sourced datasets to train the DNN Classifiers. All the QSAs are laballed by a predefined base model (The labelling method is specified in Section~\ref{modelserlectionandtraining}). After training, we conduct experiments by inferring useful new features in other 46 datasets with the classifiers.
  Before the inference, we in advance chose a test model to compare the f1-scores before and after the experiment. Finally, we analyze the efficieny and efficacy of FLFE from the experiment results.}
  \label{experimentprocedure}
\end{figure}
\indent As mentioned in section $4.2$, we collected $166$ open-sourced datasets from OpenML, $120$ of which were used to generate QSAs as training samples and the rest were for test datasets. In these datasets, the number of features varies from $3$ to $10001$, and the number of data points from hundreds and tens of thousands.
Random Forest and Logistic Regression (both with 10-fold cross-validation) are chosen as the test model. The performance improvement threshold $\theta$ is set to $1\%$. The f1-score before and after feature engineering can be calculated with the federated random forest\cite{liu2020federated}. If the f1-score shows an improvement not less than $1\%$ (same with the threshold $\theta$ in generation of training QSAs), the new feature will be directly appended into original dataset.The experimental procedure is shown in Fig.~\ref{experimentprocedure}. \\
\indent Two data augmentation techniques were also utilized. (1) Since higher threshold leads to training samples of higher quality but with fewer positive labels, we adopted SMOTE upsamling~\cite{chawla2002smote} to eliminate this imbalance. (2) Each time after labeling a feature as useful or not, we cropped the data points to generate more QSAs, as shown in Fig.\ref{crop}.\\

\subsection{The efficacy of FLFE}
\label{sec-experiment-efficacy}
\begin{figure}[htbp]
  \centering
  \includegraphics[scale=0.525]{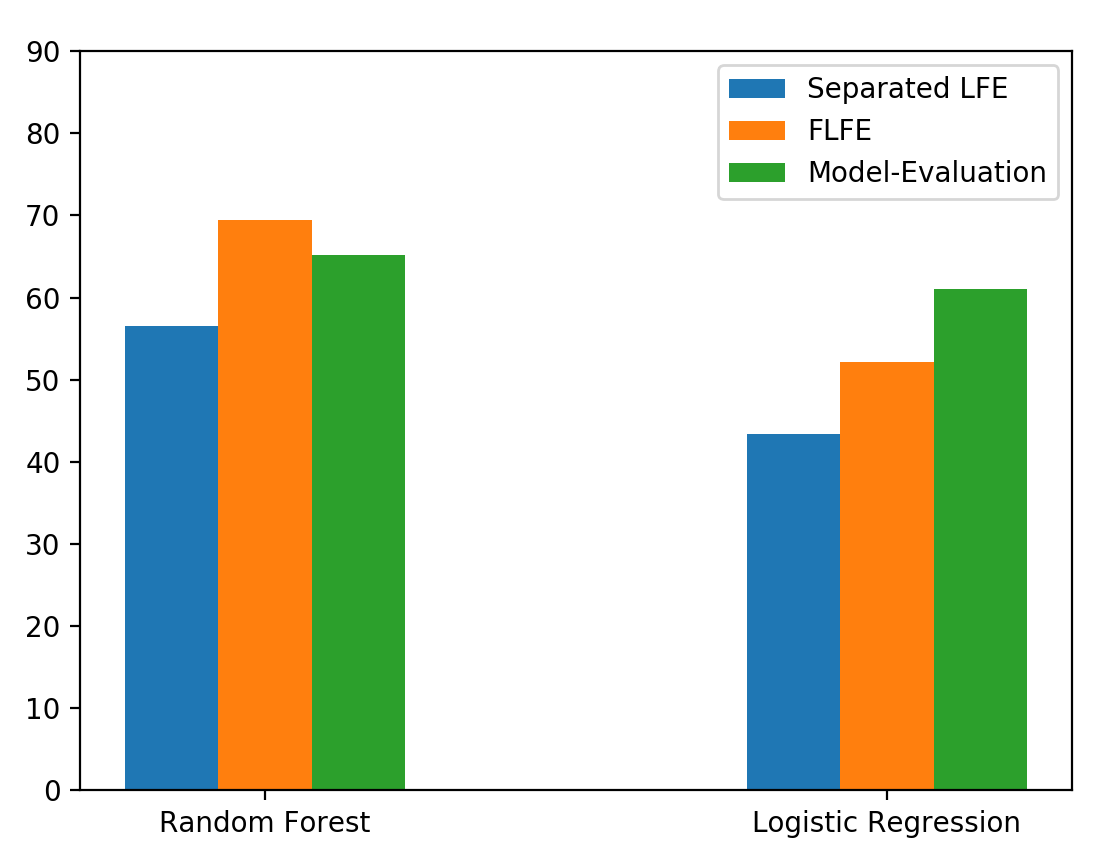}
  \caption{Number of datasets (among the 46) for which the f1-score was improved. The results show that even when the base model and the test model are inconsistent (random forest and logistic regression), FLFE can improve the f1-score. For Random Forest being the base model and test model, FLFE even improves the f1-score on more datasets than the other approaches.}
  \label{robust}
\end{figure}

\begin{table*}[t]
  \label{meaning}
  \centering
  \caption{\label{tab:test}Statistics of datasets and f1-score of FLFE and other multi-party feature engineering approaches with Random Forest being the base model. The best approach for each dataset is shown in bold. The number of added features is listed in brackets. For most datasets, the performance of FLFE was comparable to that of evaluation-based approaches, but FLFE has a far lower number of generated features.}
  \begin{tabular}{cp{2cm}<{\centering}cccc}
   \toprule
   \textbf{Dataset} & \textbf{Numerical Features} & \textbf{Bench Score} & \textbf{FLFE} & \textbf{separated LFE} & \textbf{Model Evaluation}\\
   \midrule
  churn & 16 & 81.01\% & \textbf{81.84\%}(1) & 81.68\%(1) & 81.10\%(0)\\
  boston & 16 & 78.04\% & 77.81\%(30) & \textbf{78.34\%}(1) & 77.72\%(30)\\
  statlog & 10 & 78.02\% & \textbf{81.12\%}(17) & 80.58\%(7) & 79.01\%(54)\\
  tecator & 124 & 91.75\% & \textbf{92.55\%}(102) & 92.19\%(52) & 91.07\%(568)\\
  hmeq & 11 & 87.00\% & 87.29\%(10) & 87.28\%(9) & \textbf{88.76\%}(11)\\
  triazines & 34 & 66.84\% & 70.94\%(18) & 70.52\%(11) & \textbf{71.76\%}(1065)\\
  wisconsin & 32 & 55.23\% & \textbf{56.35\%}(12) & 56.09\%(4) & 55.96\%(630)\\
  autoPrice & 15 & 90.52\% & \textbf{91.27\%}(3) & 90.39\%(0) & 90.44\%(146)\\
  clean & 168 & 92.28\% & \textbf{93.79\%}(370) & 92.91\%(129) & 93.60\%(8446)\\
  heart-statlog & 10 & 77.47\% & \textbf{81.46\%}(15) & 80.60\%(4) & 77.73\%(55)\\
  zernike & 47 & 92.81\% & 95.62\%(4) & 95.52\%(1) & \textbf{99.80\%}(171)\\
   \bottomrule
  \end{tabular}
 \end{table*}
We evaluate the efficacy of FLFE by comparing it to (1.a) Model Evaluation and (1.b) Learning Feature Engineering conducted separately on each device (Separated LFE). Model-Evaluation-based approaches preserve privacy through federated machine learning models such as Federated Forest~\cite{liu2020federated} and Federated Logistic Regression~\cite{yang2019parallel}. 
Since we only focus on the predictive performance rather than privacy, we directly apply Random Forest and Logistic Regression as test models. In Separated LFE, each device conducts unary transformations to its own features and sends the QSAs to the parameter server. \\
\indent We first chose random forest as the base model. Table 1 reports f1-scores on $11$ of the $46$ test datasets \footnote{The chosen $11$ test datasets all have names on the original features, so the generated features have clear meanings.} without feature engineering (bench score) and with feature engineering (the last $3$ columns). \\
\indent We then manifest the FLFE's robustness to the choice of test model. We used random forest and logistic regression separately as test model, while fixed random forest as the base model. Fig.~\ref{robust} shows the robustness of FLFE. 
\indent For the $2$ base models, we gauranteed that the random feature selections are the same. We generated the combinations of features for all binary transformations, and then for different base models, the same feature combinations are enumerated. \\
The experiments above show that the FLFE are comparable to Model-Evaluation-based approaches in f1-score, both superior to seperatedly LFE.
It's also worth mentioning that under tha same predictive performance, FLFE generated fewer new features than Model-Evaluation-based approaches in most cases, and thus had lower communication overhead.

\subsection{The efficiency of FLFE}
In this section, we evaluated the efficiency of FLFE, Homomorphic-Encryption-based LFE (HE-based LFE) and Model-Evaluation-based feature engineering in two aspects: running time and communication overhead. 
Experiments were done on three real-world datasets on insurance default prediction that share the same data points but with different feature domains: \textbf{Public} (30 features), \textbf{Insured Company} (248 features), and \textbf{Government} (92 features). We set up a parameter server and 3 devices, and each device keeps one of the datasets in local. 
Since the main goal in this section is to evaluate the efficiency of communication, we apply only binary transformations between input features from two different devices. 
Note that in the previous experiment in Section \ref{sec-experiment-efficacy}, we generate new features only from original features, but in this section, we generate new features also from previously generated ones. 
We believe that such a setting is closer to the real practical application of FLFE, and it more clearly shows the efficiency difference of different feature engineering approaches.
Note that since Palliar does not support encryption for division operations, we estimated the running time and communication overhead of division with those of multiplication. 

\begin{table}[htbp]
  \caption{Running time spent in each loop of different multi-party feature engineering approaches.
  Each approach makes $500$ attempts to generate useful features and the average time is reported.}
  \begin{tabular}{|l|cp{1.5cm}<{\centering}p{1.5cm}<{\centering}|}
    \hline
    \diagbox{Part}{Time}{Approach} &FLFE&HE-based LFE&Model-Evaluation\\
    \hline
    Judging& 0.0299s & 0.0611s & 1.0708s \\
    Generating& 0.0114s & 2.2933s & 0.0257s\\
    \hline
    \end{tabular}
  \end{table}

\indent  Running time consists of judging time and generating time. Judging time is spent on judging whether a transformation is useful, while generating time is for generation using transformations judged as useful. Table 2 shows that FLFE is far more superior to the other two methods in terms of running time.
The HE-based approach is much slower in generation for two reasons: (1) It needs to generate secret and public keys in every loop; (2) Encrypting and transmitting features are extremely time-consuming.
On the contrary, the bottleneck of the evaluation-based method lies in the stage of judgement, which further slows down the entire feature engineering process as the number of features increases.\\
\begin{figure}[htbp]
  \centering
  \includegraphics[scale=0.48]{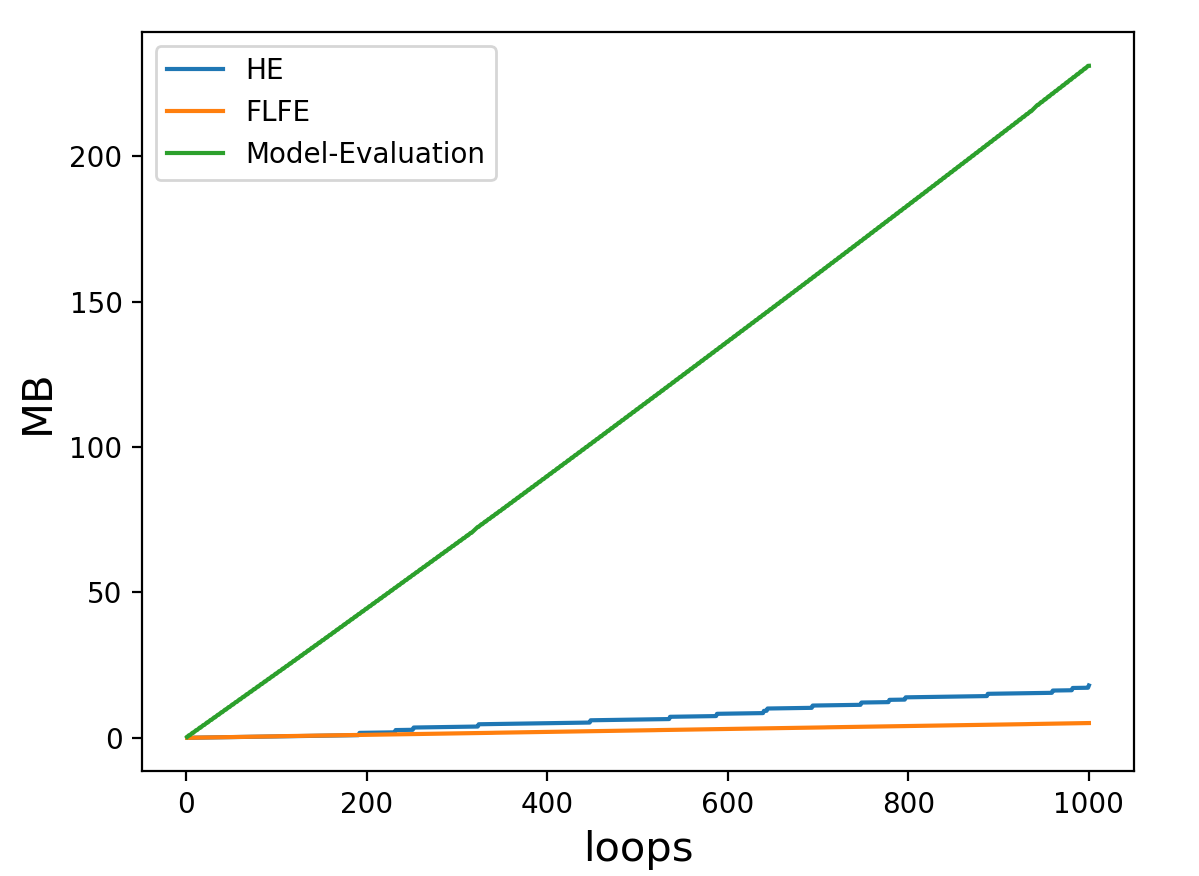}
  \caption{The trend of communication overhead of the three approaches.}
  \label{m3}
  \includegraphics[scale=0.48]{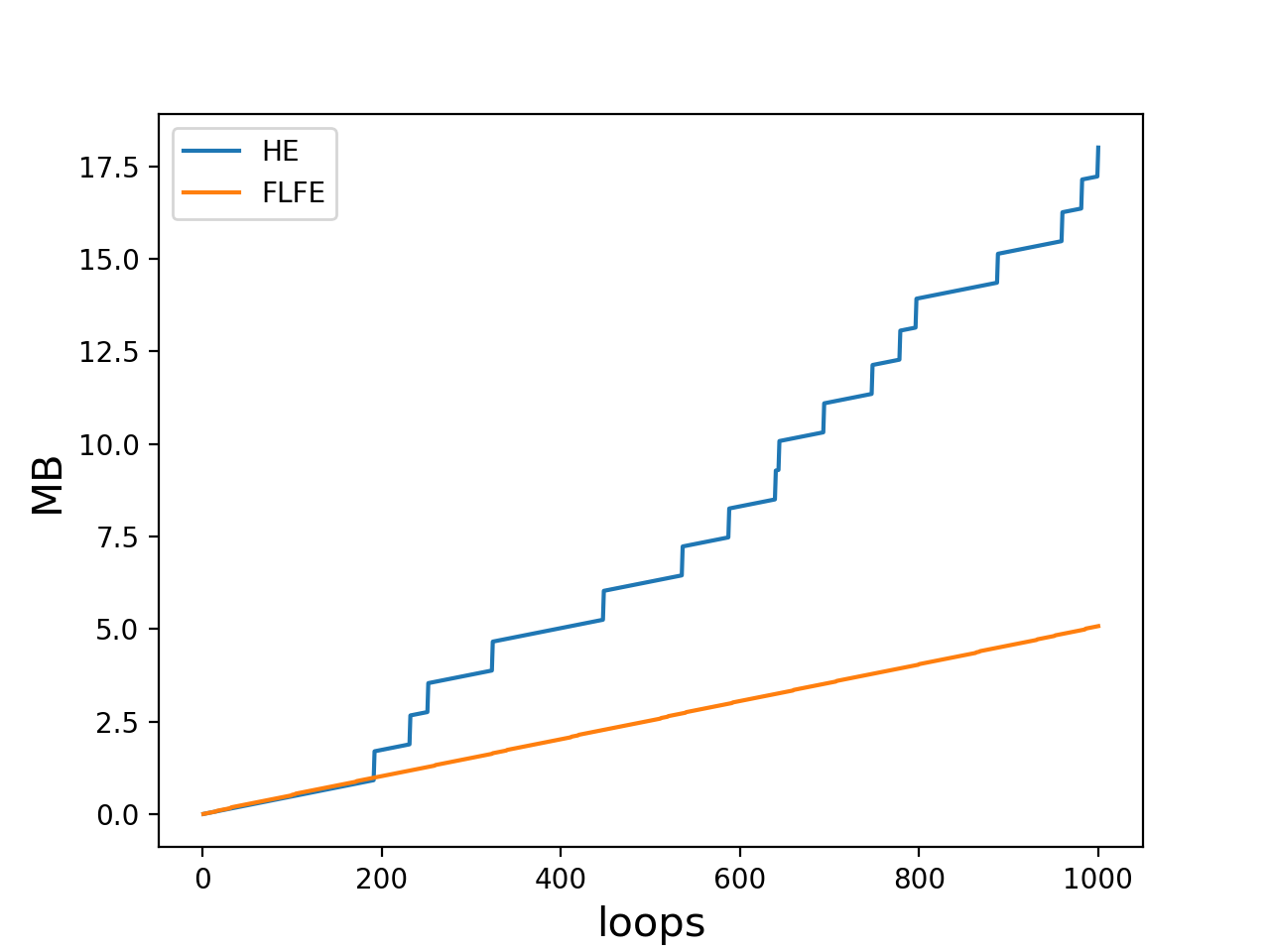}
  \caption{The trend of communication overhead under two approaches.}
  \label{m2}
\end{figure}
\indent Fig.~\ref{m3} and Fig.~\ref{m2} present us with the trend of communication overhead in multi-party feature engineering with the number of executed loops grows. Generally, the communication overhead during the execution of FLFE largely depends on 
the size of QSA. For instance, a QSA with 2 class labels and 200 bins requires a device to transmit 400 floats, namely 1600 Bytes in most computer systems. The overhead of HE-based approach in the judgement stage is roughly equal to that of FLFE.
However, homomorphically encryption of a feature makes it longer (from 8 bits to 256 bits in the experiments) and thus needs a higher bandwidth for transmission, which makes the communication overhead of HE-based approach higher. In Fig.~\ref{m2}, the communication overhead of the Evaluation-based model grows much faster than the others, and this is because for the Evaluation-based model, the feature space expands substantially as new features are added.
\section{Conclusions and Future Work}
\indent In this paper, we studied the problem of multi-party feature engineering, which poses a big challenge of privacy and communication. We proposed a framework called Federated Learning Feature Engineering (FLFE) to perform multi-party feature engineering. 
The proposed feature transmission mechanism and QSA are the two cores that make FLFE privacy-preserving and communicationally efficient. The feature transmission mechanism enables each device to keep its information private by using an erasable random mask vector
The QSA utilizes the characteristics of features, instead of the features themselves, and thus can hide the private information; Besides, QSA, being a fixed-sized array, substantially reduces communication overhead and enhance the privacy. In the setting of the same level of privacy, 
simulations showed that FLFE outperforms existing methods by a large margin in terms of communication overhead.
A future direction is to develop a method that automatically adapts the QSA size according to the dataset. 
It would also be interesting to explore more powerful representation of features, 
since QSA does not incorporate some important information such as absolute values.

\bibliographystyle{IEEEtran}
\bibliography{ref}

\begin{thebibliography}{10}
\providecommand{\url}[1]{#1}
\csname url@samestyle\endcsname
\providecommand{\newblock}{\relax}
\providecommand{\bibinfo}[2]{#2}
\providecommand{\BIBentrySTDinterwordspacing}{\spaceskip=0pt\relax}
\providecommand{\BIBentryALTinterwordstretchfactor}{4}
\providecommand{\BIBentryALTinterwordspacing}{\spaceskip=\fontdimen2\font plus
\BIBentryALTinterwordstretchfactor\fontdimen3\font minus
  \fontdimen4\font\relax}
\providecommand{\BIBforeignlanguage}[2]{{%
\expandafter\ifx\csname l@#1\endcsname\relax
\typeout{** WARNING: IEEEtran.bst: No hyphenation pattern has been}%
\typeout{** loaded for the language `#1'. Using the pattern for}%
\typeout{** the default language instead.}%
\else
\language=\csname l@#1\endcsname
\fi
#2}}
\providecommand{\BIBdecl}{\relax}
\BIBdecl

\bibitem{yang2019federated}
Q.~Yang, Y.~Liu, T.~Chen, and Y.~Tong, ``Federated machine learning: Concept
  and applications,'' 2019.

\bibitem{gentry2009fully}
C.~Gentry and D.~Boneh, \emph{A fully homomorphic encryption scheme}.\hskip 1em
  plus 0.5em minus 0.4em\relax Stanford university Stanford, 2009, vol.~20,
  no.~9.

\bibitem{chai2019secure}
D.~Chai, L.~Wang, K.~Chen, and Q.~Yang, ``Secure federated matrix
  factorization,'' 2019.

\bibitem{10.1007/11761679_29}
C.~Dwork, K.~Kenthapadi, F.~McSherry, I.~Mironov, and M.~Naor, ``Our data,
  ourselves: Privacy via distributed noise generation,'' in \emph{Advances in
  Cryptology - EUROCRYPT 2006}, S.~Vaudenay, Ed.\hskip 1em plus 0.5em minus
  0.4em\relax Berlin, Heidelberg: Springer Berlin Heidelberg, 2006, pp.
  486--503.

\bibitem{bhaskar2011noiseless}
R.~Bhaskar, A.~Bhowmick, V.~Goyal, S.~Laxman, and A.~Thakurta, ``Noiseless
  database privacy,'' in \emph{International Conference on the Theory and
  Application of Cryptology and Information Security}.\hskip 1em plus 0.5em
  minus 0.4em\relax Springer, 2011, pp. 215--232.

\bibitem{fan2010generalized}
W.~Fan, E.~Zhong, J.~Peng, O.~Verscheure, K.~Zhang, J.~Ren, R.~Yan, and
  Q.~Yang, ``Generalized and heuristic-free feature construction for improved
  accuracy,'' in \emph{Proceedings of the 2010 SIAM International Conference on
  Data Mining}.\hskip 1em plus 0.5em minus 0.4em\relax SIAM, 2010, pp.
  629--640.

\bibitem{markovitch2002feature}
S.~Markovitch and D.~Rosenstein, ``Feature generation using general constructor
  functions,'' \emph{Machine Learning}, vol.~49, no.~1, pp. 59--98, 2002.

\bibitem{dor2012strengthening}
O.~Dor and Y.~Reich, ``Strengthening learning algorithms by feature
  discovery,'' \emph{Information Sciences}, vol. 189, pp. 176--190, 2012.

\bibitem{7837936}
G.~{Katz}, E.~C.~R. {Shin}, and D.~{Song}, ``Explorekit: Automatic feature
  generation and selection,'' in \emph{2016 IEEE 16th International Conference
  on Data Mining (ICDM)}, 2016, pp. 979--984.

\bibitem{7737055}
C.~J. {Deepu}, C.~{Heng}, and Y.~{Lian}, ``A hybrid data compression scheme for
  power reduction in wireless sensors for iot,'' \emph{IEEE Transactions on
  Biomedical Circuits and Systems}, vol.~11, no.~2, pp. 245--254, 2017.

\bibitem{8249368}
B.~R. {Stojkoska} and Z.~{Nikolovski}, ``Data compression for energy efficient
  iot solutions,'' in \emph{2017 25th Telecommunication Forum (TELFOR)}, 2017,
  pp. 1--4.

\bibitem{konevcny2016federated}
J.~Kone{\v{c}}n{\`y}, H.~B. McMahan, F.~X. Yu, P.~Richt{\'a}rik, A.~T. Suresh,
  and D.~Bacon, ``Federated learning: Strategies for improving communication
  efficiency,'' \emph{arXiv preprint arXiv:1610.05492}, 2016.

\bibitem{suresh2017distributed}
A.~T. Suresh, X.~Y. Felix, S.~Kumar, and H.~B. McMahan, ``Distributed mean
  estimation with limited communication,'' in \emph{International Conference on
  Machine Learning}, 2017, pp. 3329--3337.

\bibitem{khaled2019gradient}
A.~Khaled and P.~Richt{\'a}rik, ``Gradient descent with compressed iterates,''
  \emph{arXiv preprint arXiv:1909.04716}, 2019.

\bibitem{caldas2018expanding}
S.~Caldas, J.~Kone{\v{c}}ny, H.~B. McMahan, and A.~Talwalkar, ``Expanding the
  reach of federated learning by reducing client resource requirements,''
  \emph{arXiv preprint arXiv:1812.07210}, 2018.

\bibitem{cover1999elements}
T.~M. Cover, \emph{Elements of information theory}.\hskip 1em plus 0.5em minus
  0.4em\relax John Wiley \& Sons, 1999.

\bibitem{7149287}
A.~{Ukil}, S.~{Bandyopadhyay}, and A.~{Pal}, ``Iot data compression:
  Sensor-agnostic approach,'' in \emph{2015 Data Compression Conference}, 2015,
  pp. 303--312.

\bibitem{wu2017multiscale}
X.~Wu, R.~Guo, A.~T. Suresh, S.~Kumar, D.~N. Holtmann-Rice, D.~Simcha, and
  F.~Yu, ``Multiscale quantization for fast similarity search,'' in
  \emph{Advances in Neural Information Processing Systems}, 2017, pp.
  5745--5755.

\bibitem{Wang_quantilesover}
L.~Wang, G.~Luo, K.~Yi, and G.~Cormode, ``Quantiles over data streams: An
  experimental study.''

\bibitem{kairouz2019advances}
P.~Kairouz, H.~B. McMahan, B.~Avent, A.~Bellet, M.~Bennis, A.~N. Bhagoji,
  K.~Bonawitz, Z.~Charles, G.~Cormode, R.~Cummings, R.~G.~L. D'Oliveira, S.~E.
  Rouayheb, D.~Evans, J.~Gardner, Z.~Garrett, A.~Gascón, B.~Ghazi, P.~B.
  Gibbons, M.~Gruteser, Z.~Harchaoui, C.~He, L.~He, Z.~Huo, B.~Hutchinson,
  J.~Hsu, M.~Jaggi, T.~Javidi, G.~Joshi, M.~Khodak, J.~Konečný, A.~Korolova,
  F.~Koushanfar, S.~Koyejo, T.~Lepoint, Y.~Liu, P.~Mittal, M.~Mohri, R.~Nock,
  A.~Özgür, R.~Pagh, M.~Raykova, H.~Qi, D.~Ramage, R.~Raskar, D.~Song,
  W.~Song, S.~U. Stich, Z.~Sun, A.~T. Suresh, F.~Tramèr, P.~Vepakomma,
  J.~Wang, L.~Xiong, Z.~Xu, Q.~Yang, F.~X. Yu, H.~Yu, and S.~Zhao, ``Advances
  and open problems in federated learning,'' 2019.

\bibitem{DBLP:journals/corr/McMahanMRA16}
\BIBentryALTinterwordspacing
H.~B. McMahan, E.~Moore, D.~Ramage, and B.~A. y~Arcas, ``Federated learning of
  deep networks using model averaging,'' \emph{CoRR}, vol. abs/1602.05629,
  2016. [Online]. Available: \url{http://arxiv.org/abs/1602.05629}
\BIBentrySTDinterwordspacing

\bibitem{4568207}
A.~C. {Yao}, ``How to generate and exchange secrets,'' in \emph{27th Annual
  Symposium on Foundations of Computer Science (sfcs 1986)}, 1986, pp.
  162--167.

\bibitem{mohassel2018aby3}
P.~Mohassel and P.~Rindal, ``Aby3: A mixed protocol framework for machine
  learning,'' in \emph{Proceedings of the 2018 ACM SIGSAC Conference on
  Computer and Communications Security}, 2018, pp. 35--52.

\bibitem{mohassel2017secureml}
P.~Mohassel and Y.~Zhang, ``Secureml: A system for scalable privacy-preserving
  machine learning,'' in \emph{2017 IEEE Symposium on Security and Privacy
  (SP)}.\hskip 1em plus 0.5em minus 0.4em\relax IEEE, 2017, pp. 19--38.

\bibitem{pathak2010multiparty}
M.~Pathak, S.~Rane, and B.~Raj, ``Multiparty differential privacy via
  aggregation of locally trained classifiers,'' in \emph{Advances in Neural
  Information Processing Systems}, 2010, pp. 1876--1884.

\bibitem{dwork2008differential}
C.~Dwork, ``Differential privacy: A survey of results,'' in \emph{International
  conference on theory and applications of models of computation}.\hskip 1em
  plus 0.5em minus 0.4em\relax Springer, 2008, pp. 1--19.

\bibitem{brakerski2011fully}
Z.~Brakerski and V.~Vaikuntanathan, ``Fully homomorphic encryption from
  ring-lwe and security for key dependent messages,'' in \emph{Annual
  cryptology conference}.\hskip 1em plus 0.5em minus 0.4em\relax Springer,
  2011, pp. 505--524.

\bibitem{coron2011fully}
J.-S. Coron, A.~Mandal, D.~Naccache, and M.~Tibouchi, ``Fully homomorphic
  encryption over the integers with shorter public keys,'' in \emph{Annual
  Cryptology Conference}.\hskip 1em plus 0.5em minus 0.4em\relax Springer,
  2011, pp. 487--504.

\bibitem{brakerski2014leveled}
Z.~Brakerski, C.~Gentry, and V.~Vaikuntanathan, ``(leveled) fully homomorphic
  encryption without bootstrapping,'' \emph{ACM Transactions on Computation
  Theory (TOCT)}, vol.~6, no.~3, pp. 1--36, 2014.

\bibitem{liu2019enhancing}
Z.~Liu, T.~Li, V.~Smith, and V.~Sekar, ``Enhancing the privacy of federated
  learning with sketching,'' 2019.

\bibitem{nargesian2017learning}
F.~Nargesian, H.~Samulowitz, U.~Khurana, E.~B. Khalil, and D.~S. Turaga,
  ``Learning feature engineering for classification.'' in \emph{IJCAI}, 2017,
  pp. 2529--2535.

\bibitem{rifkin2004defense}
R.~Rifkin and A.~Klautau, ``In defense of one-vs-all classification,''
  \emph{Journal of machine learning research}, vol.~5, no. Jan, pp. 101--141,
  2004.

\bibitem{nair2010rectified}
V.~Nair and G.~E. Hinton, ``Rectified linear units improve restricted boltzmann
  machines,'' in \emph{Proceedings of the 27th international conference on
  machine learning (ICML-10)}, 2010, pp. 807--814.

\bibitem{kingma2014adam}
D.~P. Kingma and J.~Ba, ``Adam: A method for stochastic optimization,''
  \emph{arXiv preprint arXiv:1412.6980}, 2014.

\bibitem{hinton2012improving}
G.~E. Hinton, N.~Srivastava, A.~Krizhevsky, I.~Sutskever, and R.~R.
  Salakhutdinov, ``Improving neural networks by preventing co-adaptation of
  feature detectors,'' \emph{arXiv preprint arXiv:1207.0580}, 2012.

\bibitem{ho2013more}
Q.~Ho, J.~Cipar, H.~Cui, S.~Lee, J.~K. Kim, P.~B. Gibbons, G.~A. Gibson,
  G.~Ganger, and E.~P. Xing, ``More effective distributed ml via a stale
  synchronous parallel parameter server,'' in \emph{Advances in neural
  information processing systems}, 2013, pp. 1223--1231.

\bibitem{liu2020federated}
Y.~Liu, Y.~Liu, Z.~Liu, Y.~Liang, C.~Meng, J.~Zhang, and Y.~Zheng, ``Federated
  forest,'' \emph{IEEE Transactions on Big Data}, 2020.

\bibitem{chawla2002smote}
N.~V. Chawla, K.~W. Bowyer, L.~O. Hall, and W.~P. Kegelmeyer, ``Smote:
  synthetic minority over-sampling technique,'' \emph{Journal of artificial
  intelligence research}, vol.~16, pp. 321--357, 2002.

\bibitem{yang2019parallel}
S.~Yang, B.~Ren, X.~Zhou, and L.~Liu, ``Parallel distributed logistic
  regression for vertical federated learning without third-party coordinator,''
  2019.

\end{thebibliography}

\end{document}